# Probabilistic Conditional Preference Networks

**Damien Bigot**
IRIT-CNRS
Toulouse University
31400 Toulouse, France
damien.bigot@irit.fr

**Hélène Fargier**
IRIT-CNRS
Toulouse University
31400 Toulouse, France
helene.fargier@irit.fr

**Jérôme Mengin**
IRIT-CNRS
Toulouse University
31400 Toulouse, France
jerome.mengin@irit.fr

**Bruno Zanuttini**
GREYC
Caen University
14000 Caen, France
bruno.zanuttini@unicaen.fr

## Abstract

This paper proposes a "probabilistic" extension of conditional preference networks as a way to compactly represent a probability distributions over preference orderings. It studies the probabilistic counterparts of the main reasoning tasks, namely dominance testing and optimisation from the algorithmical and complexity viewpoints. Efficient algorithms for tree-structured probabilistic CP-nets are given. As a by-product we obtain a linear-time algorithm for dominance testing in standard, tree-structured CP-nets.

## 1 Introduction

Modelling preferences has been an active research topic in Artificial Intelligence for more than fifteen years. In recent years, several formalisms have been proposed that are rich enough to describe complex preferences of a user in a compact way, by e.g. Rao and Georgeff [1991], Gonzales *et al.* [2008], Boutilier *et al.* [2001, 2004]. Ordinal preferences, where alternatives, or outcomes, are ranked without the use of numerical functions, are usually easier to obtain, and are the topic of this paper.

In many contexts, the preferences of the user are ill-known, e.g. because they depend on the value of non controllable state variable, or because the system has no information about the user – her preferences may then be extrapolated from information gathered for previous customers. This is typically the case in anonymous recommendation systems, where several users with similar preferences can be grouped into a single model – that can then be finely tuned to fit one particular user. In this paper, we propose to use a probability distribution over preference models to represent ill known preferences. Specifically, we propose to extend conditional preference networks (CP-nets, one of the most popular ordinal preference representation formalisms [Boutilier *et al.*, 2004]) by attaching probabilities to the local preference rules.

Probabilistic CP-nets are evoked for preference elicitation in by de Amo *et al.* [2012]. However, the authors do not give a precise semantics to their CP-nets, nor do they study their computational properties. A more general form of Probabilistic CP-net is also described by Cornelio [2012], who prove that the problem of finding the most probably optimal outcome is similar to an optimisation problem in a Bayesian network.

In the present paper, we detail the Probabilistic CP model, and especially its semantics, and provide efficient algorithms to solve the corresponding dominance and optimisation problems. After a brief presentation of CP-nets (Section 2), we present Probabilistic CP-nets, their semantics and explain how they can be used in several practical settings (Section 3). In Section 4, we give efficient algorithms and complexity results for dominance testing. In Section 5, we turn to the optimisation task and prove that it can be performed in linear time when some restriction is put on the structure of the PCP-net. Section 6 concludes the paper.

## 2 Background

We consider combinatorial objects defined over a set of $n$ variables $\mathcal{V}$. Variables are denoted by uppercase letters $A, B, X, Y, \ldots$. $\underline{X}$ denotes the domain of a variable $X$. More generally, for a set of variables $U \subseteq \mathcal{V}$, $\underline{U}$ denotes the Cartesian product of their domains. Elements of $\underline{\mathcal{V}}$ are called *objects* or *outcomes*, denoted by $o, o', \ldots$. Elements of $\underline{U}$ for some $U \subseteq \mathcal{V}$ are denoted by $u, u', \ldots$. Given two sets of variables $U \subseteq V \subseteq \mathcal{V}$ and $v \in \underline{V}$, we write $v[U]$ for the restriction of $v$ to the variables in $U$.

In this paper we essentially consider variables with a Boolean domain. We consistently write $x$ and $\bar{x}$ for the two values in the domain of $X$.

**Preference Relations** We assume that individual preferences can be represented by an order (reflexive, antisymmetric and transitive) over the set of all outcomes $\mathcal{V}$. A convenient way to specify such orders over outcomes in a multi-attribute domain is by means of *local preference rules*: each rule enables one to compare outcomes that have some specific values for some attributes. Conditional preference networks [Boutilier *et al.*, 2004] enable direct comparisons between outcomes that differ in the value of one variable only (called *swap pairs* of outcomes). Such a rule has the form $(X, u\!:\!>)$, with $X \in \mathcal{V}$, $u \in \underline{U}$ for some $U \subseteq \mathcal{V} - \{X\}$, and $>$ a total order on $\underline{X}$. According to $(X, u\!:\!>)$, for every pair of outcomes $o, o'$ such that $o[U] = o'[U] = u$ and $o[\mathcal{V} - (U \cup \{X\})] = o'[\mathcal{V} - (U \cup \{X\})]$, $o$ is preferred to $o'$ if and only if $o[X] > o[X']$. Intuitively, the rule $(X, u\!:\!>)$ can be read: "Whenever $u$ is the case, outcomes are ordered as their values for $X$ are ordered by $>$, everything else being equal".

**Example 1.** *Assuming a set of binary variables $\mathcal{V} = \{X_1, \ldots, X_4\}$, the rule $(x_3, \bar{x}_2\!:\!x_3\!>\!\bar{x}_3)$ entails that $o = x_1 \bar{x}_2 x_3 x_4$ is preferred to $o' = x_1 \bar{x}_2 \bar{x}_3 x_4$. On the other hand, it tells nothing about the preference between $o$ and $o'' = \bar{x}_1 \bar{x}_2 \bar{x}_3 x_4$ (everything else is* not *equal), nor between $x_1 x_2 x_3 x_4$ and $x_1 x_2 \bar{x}_3 x_4$ (it does not apply).*

Considering the transitive closure of the relation over swap pairs, the set of all outcomes can be (partially) ordered by a set $R$ of such rules using the notion of *flip*. An *$R$-worsening flip* is an ordered swap pair $(o, o')$ for which there is a rule $r = (X, u\!:\!>) \in R$ satisfying: $o[U] = o'[U] = u$, $o[\mathcal{V} \setminus (U \cup \{X\})] = o'[\mathcal{V} \setminus (U \cup \{X\})]$, and $o[X] > o'[X]$. A sequence of outcomes $o_1, \ldots, o_k$ is an *$R$-worsening sequence* if for $1 \leq i \leq k-1$, $(o_i, o_{i+1})$ is an $R$-worsening flip. We write $o \succ_R o'$ whenever there is an $R$-worsening sequence from $o$ to $o'$. By construction, the relation $\succ_R$ precisely captures the transitive closure of the relation induced by $R$ on swap pairs. We say that the set of rules $R$ is *consistent* if $\succ_R$ is irreflexive, and *inconsistent* otherwise.

**Conditional Preference Networks** With a conditional preference network (CP-net), one can specify preferential dependencies between variables by means of a directed graph $G = (\mathcal{V}, E)$: an edge $(X, Y)$ indicates that the preference over the domain of $Y$ may depend on the values of $X$. Given such a graph and a vertex $X \in \mathcal{V}$, we write $\mathsf{pa}(X)$ for the set of *parents* of $X$ in $(\mathcal{V}, E)$: $\mathsf{pa}(X) = \{Y \in \mathcal{V} \mid (Y, X) \in E\}$.

**Definition 1** (CP-net). *A (deterministic) CP-net $N$ over a set of variables $\mathcal{V}$ is defined by a directed graph $(\mathcal{V}, E)$, and by a* conditional preference table *for each vertex / variable $X \in \mathcal{V}$, written $\mathsf{CPT}(X)$. The table $\mathsf{CPT}(X)$ gives a local preference rule $(X, u\!:\!>)$ for each combination of values $u \in \underline{\mathsf{pa}(X)}$ for the parents of $X$.*

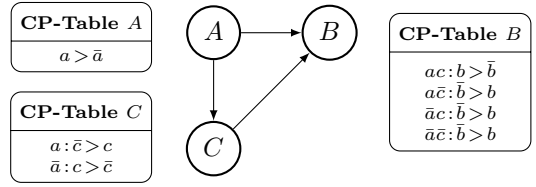

Figure 1: A deterministic CP-net

*The graph $G$ is called the* structure *of $N$.*

When $X$ is clear from the context, we write $u\!:\!>$ instead of $(X, u\!:\!>)$ for a conditional rule. For instance, given a CP-net and a binary variable $B$ with a single parent $A$, we write $a\!:\!b\!>\!\bar{b}$ for the rule $(B, a\!:\!b\!>\!\bar{b})$. We also write $>^u_{N,X}$ for the total order over $\underline{X}$ specified by a CP-net $N$ for some variable $X$ and some assignment $u \in \underline{\mathsf{pa}(X)}$. Finally, if no ambiguity can arise, we use the same notation for a CP-net and its set of local preference rules. In particular, we write $o \succ_N o'$ to indicate that there is a worsening sequence from $o$ to $o'$ using the rules of $N$. When this is the case, we also say that $N$ *entails* $o \succ o'$.

For complexity analysis, we write $|N|$ for the size of $N$, defined to be the number of symbols needed to write all rules, where writing a rule $(X, u\!:\!>)$ is considered to require $|U| + |\underline{X}|$ symbols. We also use specific classes of CP-nets, defined by restrictions on their structure $G$. For instance, the class of *acyclic* (resp. tree-structured) CP-nets is the class of CP-nets whose structure is an acyclic graph (resp. a forest).

A CP-net $N$ is said to be *inconsistent* if the set of rules of $N$ is inconsistent, and *consistent* otherwise. It is known [Boutilier *et al.*, 2004] that all acyclic CP-nets are consistent, but the converse is not true in general.

**Example 2.** *Figure 1 shows a CP-net over three variables $A, B, C$. This CP-net is consistent (it is acyclic), and it entails $abc \succ \bar{a}\bar{b}c$, as can be seen from the worsening sequence $abc \succ a\bar{b}c \succ \bar{a}\bar{b}c$, which uses the first rule in $\mathsf{CPT}(B)$, then the rule on $A$.*

Given a (consistent) CP-net $N$ the two main reasoning problems are *dominance* and *optimisation*. Dominance is the problem of deciding whether $N$ entails $o \succ o'$ for two given outcomes $o, o'$, and optimisation consists in computing the "best" outcomes according to $N$; that is, the outcomes which are undominated under $\succ_N$. For acyclic CP-nets, optimisation is feasible in linear time, and there is always a unique optimal outcome. Contrastingly, testing dominance is PSPACE-complete for unrestricted CP-nets, NP-hard for acyclic ones, and quadratic for tree-structured ones [Goldsmith *et al.*, 2005].

## 3 Probabilistic CP-Nets

When the preferences of the user are ill-known, typically because they depend on the value of non controllable state variables, or because the system has few information about the user, we would like to be able to answer questions like "What is the probability that $o$ is preferred to $o'$ by some unknown agent?". *Probabilistic CP-nets* enable to compactly represent a probability distribution over CP-nets and answer such queries. A typical application is recommendation, where the preferences of the current (anonymous) user are extrapolated from profiles or from information gathered from previous customers in order to estimate how likely it is that a new customer makes a given choice.

**Definition 2** (PCP-net). *A probabilistic conditional preference network $\mathcal{N}$, or PCP-net, over a set of variables $\mathcal{V}$, is defined by a directed graph $G=(\mathcal{V}, E)$ and, for each vertex / variable $X \in \mathcal{V}$, a probabilistic conditional preference table, written* PCPT$(X)$. *The PCP-table on $X$ gives, for each assignment $u \in \mathsf{pa}(X)$, a probability distribution over the set of the total orders on $\underline{X}$. We write $p_{\mathcal{N},X}^u$ for this distribution. We also call $G$ the* structure *of $\mathcal{N}$.*

In particular, when all variables are binary, a PCP-table on $X$ gives for each assignment $u \in \mathsf{pa}(X)$ a probability distribution on the set of two orders $\{x > \bar{x}, \bar{x} > x\}$[1]. For brevity, we write $u : x > \bar{x}$ $(p)$ for the distribution which assigns probability $p$ to $u : x > \bar{x}$ and $1-p$ to $u : \bar{x} > x$, as in Figure 2.

Taken as a whole, a PCP-net $\mathcal{N}$ is not intended to represent a preference relation. Rather, it represents a probability distribution over a set of (deterministic) CP-nets, namely, those which are compatible with $\mathcal{N}$.

**Definition 3** (compatibility, probability). *A (deterministic) CP-net $N$ is said to be* compatible *with a PCP-net $\mathcal{N}$, or to be $\mathcal{N}$-compatible, if it has the same structure as $\mathcal{N}$. In this case we write $N \propto \mathcal{N}$. If $N$ is $\mathcal{N}$-compatible, we define the* probability of $N$ *according to $\mathcal{N}$ by $p_{\mathcal{N}}(N) = \prod_{X \in V, u \in \underline{\mathsf{pa}(X)}} p_{\mathcal{N},X}^u (>_{N,X}^u)$.*

It easily comes that $p_{\mathcal{N}}$ is a probability distribution over the set of deterministic $\mathcal{N}$-compatible CP-nets.

**Example 3.** *Figure 2 shows a PCP-net $\mathcal{N}$ over variables $X, Y, Z, T, U, V$. The first rule on $Y$, for instance, says that there is a .2 probability that a de-*

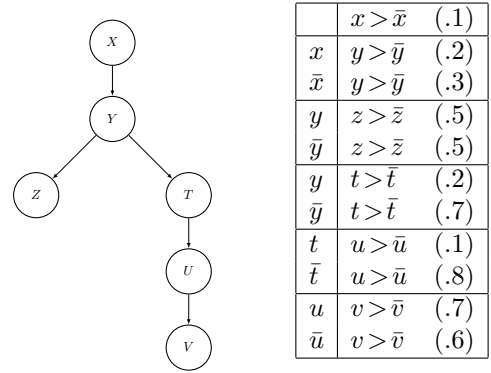

Figure 2: A probabilistic CP-net

*terministic CP-net drawn at random contains the rule $x : y > \bar{y}$; otherwise (i.e. with probability $1 - 0.2$) it contains the opposite rule $x : \bar{y} > y$. Independently, there is a .3 probability that it contains $\bar{x} : y > \bar{y}$. In particular, there is a $.2 \times .3 = .06$ probability that it contains both and hence, that $y$ is unconditionally preferred to $\bar{y}$.*

*The deterministic, $\mathcal{N}$-compatible CP-net with the negated value of each variable always preferred has probability $p = (1-.1) \times (1-.2) \times \cdots \times (1-.7) \times (1-.6)$.*

Observe that when $\mathcal{N}$ contains cycles, $p_{\mathcal{N}}$ may be nonzero for some inconsistent CP-nets, which seems undesirable. Moreover, while deciding whether a given (cyclic) CP-net is consistent is a PSPACE-hard problem [Goldsmith *et al.*, 2005], the task is tractable in the acyclic case. Therefore, the remainder of the paper considers acyclic structures only.

**Motivation** Our motivation for studying PCP-nets stems from several different applicative settings. In the first one, the preference of the current (anonymous) user are unknown but the system has at its disposal the preferences of each of $m$ individuals (e.g., past customers), and for each one the preferences are given by a (deterministic) CP-net $N_i$ over some common structure $G$. Then the probabilistic CP-net $\mathcal{N}$ over the graph $G$ defined by $p_{\mathcal{N},X}^u(>) = \#\{i \,|\, (X, u : >) \in N_i\}/m$ (proportion of $N_i$'s which contains this rule, independently from other rules) provides a compact summary of the set of all individual preferences.

Such aggregation obviously induces a certain approximation of the distribution of preferences in the population. Namely, the probability of a given CP-net $N$ as computed from the PCP-net $\mathcal{N}$ (Definition 3) is in general different from the proportion of individuals which indeed have the preferences encoded by $N$. Precisely, the construction amounts to approximate the distribution of preference relations as an independent one, considering each rule as a random variable. This may look like a crude approximation; still, as shown below, it is sound and complete for some restricted queries. Moreover, we discuss in Section 6 how PCP-nets can be extended to richer representations of distributions.

---
[1] The situation might become less simple when the size of the domains increases: the PCP-table on $X$ gives for each assignment $u \in \mathsf{pa}(X)$ a probability distribution on the possible orders on $\underline{X}$. Allthrough there are $|\underline{X}|!$ potential orders, only a few may receive a significant probability and shall explicity appear in the PCP table; the remaining mass of probabilty is then assumed to be shared by the other, less significant, preference orders.

A close setting in which PCP-nets may prove useful is the one where a system interacts with a lot of individuals, but each one gives only a few preferences. For instance, in a recommender system, assume that each customer implicitly gives a preference of the form $u\!:\!x\!>\!\bar{x}$, by choosing one of two objects in a swap pair. This is the case when, say, a customer chooses the colour for a car in a context of interactive configuration [Gelle and Weigel, 1996]: she implicitly expresses a preference of the form $u\!:\!x\!>\!\bar{x}$, where $U$ is the set of variables that have already been assigned and $x$ is the chosen colour. Individual (deterministic) rules are thus obtained from different customers and, in the absence of other information, it clearly makes sense to aggregate these rules independently from each other.

A third applicative context is one in which only one person or agent expresses her preferences, but some noise must be taken into account, due to the elicitation process, or possibly from the person's preferences themselves (*e.g.*, "for dinner, with pasta bolognese I *most often* prefer having parmiggiano"). Assuming independent noise on each rule, PCP-nets are well suited for representing such preferences (through a rule like: dinner∧bolognese:parmiggiano> ¬parmiggiano(.9) for the above example).

In all these settings, a PCP-net comes with a structure, which constrains the dependencies among variables. In case several CP-nets are aggregated into a PCP-net, it is natural to build the latter with the union of all individual graphs as its own structure. Indeed, an individual CP-net with structure $(V, E)$ can always be seen as one over $(V, E')$, for any superset $E'$ of $E$. In the remainder of this paper we will mainly focus on *tree-structured* (P)CP-nets. While this is a clear restriction on expressivity, as we will see even such networks raise nontrivial computational problems.

**Reasoning Tasks** Since a PCP-net represents a probability distribution on a set of deterministic CP-nets, the most natural queries are the following.

**Definition 4** (probability of dominance). *Given a PCP-net $\mathcal{N}$ and two outcomes $o, o'$, the probability of $o \succ o'$, written $p_\mathcal{N}(o \succ o')$, is defined to be the probability mass of $\mathcal{N}$-compatible CP-nets which entail $o \succ o'$:*

$$p_\mathcal{N}(o \succ o') = \sum_{N \propto \mathcal{N}, o \succ_N o'} p_\mathcal{N}(N)$$

Clearly enough, the probability of $o \succ o'$ given $\mathcal{N}$ is precisely the probability, when drawing a CP-net at random according to $p_\mathcal{N}$, of obtaining one which entails $o \succ o'$. In the remainder of the paper, we will essentially study how to compute such probability.

The second query is the probabilistic counterpart of optimisation in deterministic CP-nets.

**Definition 5** (probability of being optimal). *Given an acyclic PCP-net $\mathcal{N}$ and an outcome $o$, the* probability *for $o$ to be optimal, written $p_\mathcal{N}(o)$, is defined to be the probability mass of $\mathcal{N}$-compatible CP-nets which have $o$ as their optimal outcome* [2].

Interestingly, despite the important approximation induced when summarising a population of CP-nets into a single PCP-net, some reasoning tasks can be performed exactly with the approximation (PCP-net) only. So let $\mathcal{N}$ be an acyclic PCP-net built from the rulewise aggregation of individual CP-nets.

**Proposition 1.** *Let $\mathcal{N}$ be an acyclic PCP-net and $\{o, o'\}$ a swap pair of outcomes, differing only on the value of $X$. The probability $p_\mathcal{N}(o \succ o')$ is precisely the proportion of individual CP-nets which entail $o \succ o'$.*

*Proof.* This follows from the fact that for acyclic $G$, a deterministic CP-net $N$ entails $o \succ o'$ if and only if it contains the rule $o[\mathsf{pa}(X)]\!:\!o[X]\!>\!\bar{o}[X]$ [Koriche and Zanuttini, 2009, Lemma 1]. □

Another interesting property is the preservation of local Condorcet winners [Xia *et al.*, 2008, Li *et al.*, 2011], also called "hypercubewise Condorcet winners" by Conitzer *et al.* [2011]: they are the outcomes $o$ which are preferred by at least one half of the individual CP-nets to all $o'$ that differ from $o$ in the value of one variable only. Proposition 1 proves that the hypercubewise Condorcet winners are the outcomes that dominate each of their neighbors in the aggregated PCP-net with a probability of at least 0.5.

Moreover, let us insist that PCP-nets may serve other purposes than preference aggregation, as, for instance, modelling ill-known preferences of a single user, and that in such settings no approximation occurs.

## 4 Complexity of Dominance Testing

We now study the complexity of the dominance problem, namely, of computing the probability of $o \succ o'$ given a PCP-net $\mathcal{N}$. We restrict our attention to tree-structured CP-nets, that is, to the case when $G$ is acyclic and assigns at most one parent to each variable. This arguably cannot capture all interesting dependency structures among variables, but as we will see this is already a nontrivial setting.

---

[2]Under our assumption of acyclicity, each CP-net is guaranteed to have a unique optimal outcome, hence the soundness of the definition.

We first give a generic construction, and use it for deriving a fixed-parameter tractability result, with the number of variables over which $o, o'$ differ as the parameter. Then as a by-product, we derive an interesting result for *deterministic* CP-nets, namely, an $O(n)$ algorithm for dominance testing. Finally, we show that with slightly more general structures, computing the probability of dominance is #P-hard.

### 4.1 Construction

The cornerstone of our results is a characterisation of all deterministic CP-nets for which there exists a worsening sequence from $o$ to $o'$, given a tree structure $G$. The characterisation is given as a propositional formula for each leaf $X$, written $\mathsf{worsen}^{o,o'}(X)$, over variables of the form $y{:}x{>}\bar{x}$, $\bar{y}{:}x{>}\bar{x}$, etc., with $Y$ the parent of $X$ in $G$. An assignment of, say, $y{:}x{>}\bar{x}$ to $\top$, means that the corresponding CP-net contains the rule; a complete assignment to all variables thus defines a deterministic CP-net with structure $G$.

Precisely, fix a forest $G = (V, E)$ and two outcomes $o, o'$. For each variable $X$ with no parent in $G$, we introduce the propositional variable $x{>}\bar{x}$, and we write $\bar{x}{>}x$ for its negation (because $>$ is total, $x{>}\bar{x}$ is true iff $\bar{x}{>}x$ is false). Similarly, for each variable $X$ with $\mathsf{pa}(X) = \{Y\}$, and $y^\epsilon \in \{y, \bar{y}\}$, we introduce the propositional variables $y{:}x{>}\bar{x}$ and $\bar{y}{:}x{>}\bar{x}$ and we write $y{:}\bar{x}{>}x$ and $\bar{y}{:}\bar{x}{>}x$ for their respective negations.

Boutilier *et al.* [2004, Appendix A] show that a worsening sequence may include up to $\Theta(n)$ changes of the value of some variable, even with binary tree-structured CP-nets. We exploit it by reasoning on the number of changes of each variable.

Precisely, the formula $\mathsf{change}_k^{o,o'}(X)$ means that there is a worsening sequence in which $X$ alternates value at least $k$ times, starting from its value in $o$ and ending with its value in $o'$. For instance, $\mathsf{change}_3^{x \dots, \bar{x} \dots}(X)$ means that there is a worsening sequence in which $X$ successively takes values $x, \bar{x}, x, \bar{x}$ (at least 4 values and 3 alternations). Formula $\mathsf{change}_k^{o,o'}(X)$ is defined inductively in Table 1, where $Y$ denotes the parent of $X$. We give the formulas for the case where $o[X] = x, o[Y] = y$, the other cases can be obtained by symetry. Then $\mathsf{worsen}^{o,o'}(X)$ is defined as follows:

- $\mathsf{worsen}^{o,o'}(X) = \mathsf{change}_0^{o,o'}(X)$ if $o[X] = o'[X]$;
- $\mathsf{worsen}^{o,o'}(X) = \mathsf{change}_1^{o,o'}(X)$ otherwise.

**Example 4.** *Consider again the PCP-net depicted in Figure 2, and let $o = xyztuv, o' = \bar{x}y\bar{z}t\bar{u}v$. The corresponding formulas are given in Table 2.*

In the following, we write $o[\geq X]$ for $o$ restricted to the variables which are ascendants of $X$ in $G$ ($X$ included).

**Proposition 2.** *There is a worsening sequence from $o[\geq X]$ to $o'[\geq X]$ in which $X$ changes value at least $k$ times if and only if $N$ is a model of $\mathsf{change}_k^{o,o'}(X)$.*

*Proof.* The proof goes by induction on the definition of the formula. For lack of space, we omit the proof for the base cases.

For the inductive step, we give a proof only for Rule 1 (Case $o[X] = o'[X] = x, o[Y] = o'[Y] = y$). The other rules are proved in exactly the same manner. So assume first that $N$ satisfies the formula in Rule 1. Then by IH there is a worsening sequence

$$\omega_1 y, \omega_2 \bar{y}, \dots, \omega_k \bar{y}, \omega_{k+1} y$$

in which all $\omega_i$'s are assignments to the proper ascendants of $Y$ and $\omega_1 y$ (resp. $\omega_{k+1} y$) is $o[\geq Y]$ (resp. $o'[\geq Y]$). If moreover $N$ satisfies the first disjunct $(y{:}x{>}\bar{x} \wedge \bar{y}{:}\bar{x}{>}x)$, since the value of $X$ has no influence on the preference over the values of $Y$ we can build the sequence

$$\omega_1 yx, \omega_1 y\bar{x}, \omega_2 \bar{y}\bar{x}, \omega_2 \bar{y}x, \dots, \omega_k \bar{y}\bar{x}, \omega_k \bar{y}x, \omega_{k+1} yx$$

which is a worsening sequence from $o[\geq X]$ to $o'[\geq X]$ where $X$ changes value $k$ times, as desired. Similarly, if $N$ satisfies the second disjunct, we can build the following sequence, where $X$ also changes value $k$ times.

$$\omega_1 yx, \omega_2 \bar{y}x, \omega_2 \bar{y}\bar{x}, \dots, \omega_k \bar{y}x, \omega_k \bar{y}\bar{x}, \omega_{k+1} y\bar{x}, \omega_{k+1} yx$$

Conversely, we show that if there is a sequence as in the claim, then $N$ satisfies the formula in Rule 1. Let

$$\omega_1 x, \omega_2 \bar{x}, \dots, \omega_k \bar{x}, \omega_{k+1} x$$

be a sequence from $o[\geq X]$ to $o'[\geq X]$ in which $x$ changes value at least $k \geq 2$ times. There must be two opposite rules on $X$, for otherwise $X$ cannot change value back and forth. Hence the disjunction in the definition of $\mathsf{change}_k^{o,o'}(X)$ is satisfied. Moreover, these rules must fire alternatively at least $k$ times overall, hence $Y$ must take at least $k$ different values in the sequence $\omega_1, \omega_2, \dots, \omega_{k+1}$, that is, change value at least $k - 1$ times. But since it starts and ends with the same value $y$ and $k - 1$ is odd, in fact it must change at least $k$ times. Hence by IH, $N$ must satisfy $\mathsf{change}_k^{o,o'}(Y)$. $\square$

**Proposition 3.** *There is a worsening sequence from $o$ to $o'$ if and only if $N$ satisfies the formula $\bigwedge_X \mathsf{worsen}^{o,o'}(X)$, where $X$ ranges over all leaves in the tree structure of $N$.*

*Proof.* Proposition 2 shows the claim if $G$ is reduced to a chain. For the more general setting, consider two branches with a common part above $X$ (included), and

$$\text{worsen}^{o,o'}(V) = \text{worsen}^{o,o'}(U)$$
$$\text{worsen}^{o,o'}(U) = (t{:}u>\bar{u} \wedge \text{change}_0^{o,o'}(T)) \vee (\bar{t}{:}u>\bar{u} \wedge \text{change}_2^{o,o'}(T))$$
$$\text{change}_0^{o,o'}(T) = \text{worsen}^{o,o'}(Y)$$
$$\text{change}_2^{o,o'}(T) = ((y{:}t>\bar{t} \wedge \bar{y}{:}\bar{t}>t) \vee (y{:}\bar{t}>t \wedge \bar{y}{:}t>\bar{t})) \wedge \text{change}_2^{o,o'}(Y)$$
$$\text{worsen}^{o,o'}(Z) = (y{:}z>\bar{z} \wedge \text{change}_0^{o,o'}(Y)) \vee (\bar{y}{:}z>\bar{z} \wedge \text{change}_2^{o,o'}(Y))$$
$$\text{worsen}^{o,o'}(Y) = \text{change}_0^{o,o'}(Y)$$
$$\text{change}_0^{o,o'}(Y) = \text{worsen}^{o,o'}(X)$$
$$\text{change}_2^{o,o'}(Y) = (x{:}y>\bar{y} \wedge \bar{x}{:}\bar{y}>y \wedge \text{change}_1^{o,o'}(X)) \vee (x{:}\bar{y}>y \wedge \bar{x}{:}y>\bar{y} \wedge \text{change}_3^{o,o'}(X))$$
$$\text{change}_1^{o,o'}(X) = x>\bar{x}$$
$$\text{change}_3^{o,o'}(X) = \bot$$

| Base cases ($\text{Pa}(X)=\emptyset$ or $k\leq 1$) | | | |
|---|---|---|---|
| $\mathbf{Pa(X)}$ | $\mathbf{o, o'}$ | $\mathbf{k}$ | $\text{change}_k^{o,o'}(X)$ |
| $\emptyset$ | $o[X]=o'[X]$ | $0$ | $\top$ |
| $\emptyset$ | $o[X]=o'[X]$ | $>0$ | $\bot$ |
| $\emptyset$ | $o[X]=x, o'[X]=\bar{x}$ | $0$ | $\text{change}_1^{o,o'}(X)$ |
| $\emptyset$ | $o[X]=x, o'[X]=\bar{x}$ | $1$ | $x>\bar{x}$ |
| $\emptyset$ | $o[X]=x, o'[X]=\bar{x}$ | $>1$ | $\bot$ |
| $\{Y\}$ | $o[X]=o'[X]$ | $0$ | $\text{change}_0^{o,o'}(Y)$ |
| $\{Y\}$ | $o[X]\neq o'[X]$ | $0$ | $\text{change}_1^{o,o'}(X)$ |
| $\{Y\}$ | $o[X]=o'[X]$ | $1$ | $\text{change}_2^{o,o'}(X)$ |
| $\{Y\}$ | $o[X]=x, o'[X]=\bar{x}, o[Y]=o'[Y]=y$ | $1$ | $(y{:}x>\bar{x} \wedge \text{change}_0^{o,o'}(Y)) \vee (\bar{y}{:}x>\bar{x} \wedge \text{change}_2^{o,o'}(Y))$ |
| $\{Y\}$ | $o[X]=x, o'[X]=\bar{x}, o[Y]=y, o'[Y]=\bar{y}$ | $1$ | $(y{:}x>\bar{x} \vee \bar{y}{:}x>\bar{x}) \wedge \text{change}_1^{o,o'}(Y)$ |

| Inductive step ($\text{Pa}(X)\neq\emptyset$ and $k>1$) | | | |
|---|---|---|---|
| Rule | k | $\mathbf{o[X], o'[X], o[Y], o'[Y]}$ | $\text{change}_k^{o,o'}(X)$ |
| 0 | odd | $x, x,$ indifferent, indiff. | $\text{change}_{k+1}^{o,o'}(X)$ |
| 1 | even | $x, x, y, y$ | $((y{:}x>\bar{x} \wedge \bar{y}{:}\bar{x}>x) \vee (y{:}\bar{x}>x \wedge \bar{y}{:}x>\bar{x})) \wedge \text{change}_k^{o,o'}(Y)$ |
| 2 | even | $x, x, y, \bar{y}$ | $(y{:}x>\bar{x} \wedge \bar{y}{:}\bar{x}>x \wedge \text{change}_{k-1}^{o,o'}(Y)) \vee (y{:}\bar{x}>x \wedge \bar{y}{:}x>\bar{x} \wedge \text{change}_{k+1}^{o,o'}(Y))$ |
| 3 | even | $x, \bar{x},$ indifferent, indiff. | $\text{change}_{k+1}^{o,o'}(X)$ |
| 4 | odd | $x, \bar{x}, y, y$ | $(y{:}x>\bar{x} \wedge \bar{y}{:}\bar{x}>x \wedge \text{change}_{k-1}^{o,o'}(Y)) \vee (y{:}\bar{x}>x \wedge \bar{y}{:}x>\bar{x} \wedge \text{change}_{k+1}^{o,o'}(Y))$ |
| 5 | odd | $x, \bar{x}, y, \bar{y}$ | $((y{:}x>\bar{x} \wedge \bar{y}{:}\bar{x}>x) \vee (y{:}\bar{x}>x \wedge \bar{y}{:}x>\bar{x})) \wedge \text{change}_k^{o,o'}(Y)$ |

Table 1: Inductive definition of the formula $\text{change}_k^{o,o'}(X)$

Table 2: Formulas for the example of Figure 2 with $o=xyztuv, o'=\bar{x}y\bar{z}t\bar{u}v$

write $\mathcal{X}, \mathcal{Y}, \mathcal{Z}$ for the set of variables above $X$ (included), in the left subtree below $X$, and in the right subtree below $X$, respectively.

Clearly, if there is a worsening sequence from $o$ to $o'$, then $N$ must satisfy the formula for both branches (by Proposition 2). For the converse, if $N$ satisfies both formulas, by Proposition 2 again there is a worsening sequence from the outcome $o[\geq Y]=o[\mathcal{X}]o[\mathcal{Y}]$ to $o'[\geq Y]=o'[\mathcal{X}]o'[\mathcal{Y}]$, and one from $o[\mathcal{X}]o[\mathcal{Z}]$ to $o'[\mathcal{X}]o'[\mathcal{Z}]$. By construction of the formula $\text{worsen}^{o,o'}(\cdot)$, there is one of these sequences in which the values of the variables above $\mathcal{X}$ change most, say, the one for $\mathcal{Y}$. Then since $\mathcal{Y}$ and $\mathcal{Z}$ are independent of each other, all flips over $\mathcal{Z}$ can also be performed in this sequence and interleaved with those over $\mathcal{Y}$. In this manner we get a worsening sequence from $o$ to $o'$, as desired.

The proof for a generic forest is obtained by applying this reasoning inductively on the set of branches. □

### 4.2 Efficient Dominance Testing

From Propositions 2–3 we first derive a *fixed-parameter tractable* (FPT) algorithm for dominance testing in tree-structured PCP-nets. Recall that a FPT algorithm is one with running time $O(f(k).n^c)$, where $n$ is the size of the input, $c$ is a constant, $f$ is a computable function, and $k$ is some measure of the input size, called the *parameter* and assumed to be small [Flum and Grohe, 2006]. The running time of such an algorithm is essentially a polynomial modulo a factor which may be exponential (or more) in the value of the parameter.

As a parameter for the dominance problem in PCP-nets, we take the number of variables which have a different value in $o$ and $o'$. This makes sense in practice since typically, in applications, one does not have to compare objects which are completely different from each other. For instance, in recommender systems a recommendation is likely to take place once the customer has fixed a number of features of the product which she wants to buy (*e.g.*, "I want a recent Blues album, cheaper than such price, etc.").

**Definition 6.** *The* parameterized dominance problem for tree-structured PCP-nets, *written* `p-Tree-PDominance`, *is defined by:*
  Input:       a tree-structured PCP-net $\mathcal{N}$, $o$, $o'$
  Parameter:   $k = |\{X \in \mathcal{V} \mid o[X] \neq o'[X]\}|$
  Output:      the probability of $o \succ o'$ according to $\mathcal{N}$

**Theorem 1.** *The problem* `p-Tree-Dominance` *is fixed-parameter tractable. Precisely, it admits an algorithm with running time in $O(2^{2k^2} n)$.*

*Proof.* For each leaf variable $X$ in the tree of $\mathcal{N}$, the algorithm first unrolls the formula $\mathsf{worsen}^{o,o'}(X)$. Each time if finds two different recursive calls (*e.g.*, on $k-1$ and $k+1$ in the second rule), it splits the formula into two parts. By construction the algorithm ends up with

$$\Phi^X = \{\varphi_1^X, \varphi_2^X, \ldots, \varphi_{n_X}^X\}.$$

The $\varphi_i^X$ are mutually inconsistent, since the recursive calls in each rule are conditioned on mutually inconsistent formulas about the current node. Moreover, by Proposition 2, a CP-net $N \propto \mathcal{N}$ satisfies $o[\geq X] \succ o'[\geq X]$ if and only if it satisfies one of these formulas.

Now define $\Phi$ to be the set of formulas

$$\Phi = \{\varphi^{X_1} \wedge \varphi^{X_2} \wedge \cdots \wedge \varphi^{X_k} \mid X_i \text{ a leaf}, \varphi^{X_i} \in \Phi^{X_i}\}$$

that is, the "cartesian products" of the $\Phi^X$'s (over all leaves). By construction, the conjunctions in $\Phi$ are mutually inconsistent, and a CP-net $N \propto \mathcal{N}$ satisfies $o \succ o'$ if and only if it satisfies one of them (Proposition 3). It follows that the probability sought for can be computed in time $O(|\Phi| \cdot n)$: the weight of each conjunction of $\Phi$ can be obtained by multiplying the probabilities of the corresponding rules in $\mathcal{N}$, in time $O(n)$, and by mutual inconsistency the result is obtained by summing up over the elements of $\Phi$. Observe that some elements of $\Phi$ may be inconsistent formulas, but this can be detected efficiently since by construction they are conjunctions of literals.

To complete the proof we only need to bound the size of $\Phi$. First consider $\Phi^X$ for some variable $X$: by construction, $|\Phi^X|$ is $2^\ell$, where $\ell$ is the number of rules used which result in two different recursive calls. This is the case for the second and third rules of Table 1 only, that is, when exactly one of $X, Y$ has a different value in $o$ and $o'$. It follows $\ell \leq 2k$, hence $|\Phi^X| \leq 2^{2k}$ for all $X$ and finally, $|\Phi| \leq (2^{2k})^k = 2^{2k^2}$, as claimed. □

### 4.3 The Deterministic Case

As an interesting by-product, we now derive a linear-time algorithm for dominance testing in tree-structured *deterministic* CP-nets. This improves on the quadratic running time of the TreeDT algorithm of Boutilier *et al.* [2004], and may seem odd since the smallest worsening sequence may be of quadratic size. This is actually not contradictory: our result says that it is possible to decide whether this sequence exists, without explicitly constructing it.

**Theorem 2.** *The dominance problem for tree-structured (deterministic) CP-nets on $n$ variables can be solved in linear time $O(n)$.*

*Proof.* The algorithm[3] simply consists of deciding whether $N$ satisfies the formula $\bigwedge_X \mathsf{worsen}^{o,o'}(X)$, where $X$ ranges over all leaves in the structure of $N$. This can be done efficiently because for all four general rules, $N$ necessarily satisfies at most one of the two disjuncts and hence, only one recursive call is involved at each step. The only point to be checked is that the algorithm can avoid considering the same variable several times along different branches.

To do so, the algorithm unrolls the formulas $\mathsf{worsen}^{o,o'}(X)$ in parallel. Each time two branches meet at a node $X$, this must be through recursive calls fired by the children of $X$. By construction, these calls are all of the form $\mathsf{change}_{k_i}^{o,o'}(X)$, and by construction and Proposition 3, all of them must be satisfied.

---

[3]This proof is a direct application of our PCP-algorithm to the deterministic case; but the query may be addressed by more dedicated and simpler (linear) algorithms. We thank an anonymous reviewer for pointing this to us.

Recall that $\mathsf{change}_{k_i}^{o,o'}(X)$ reads "$X$ changes value *at least* $k_i$ *times*". Then the algorithm simply needs to replace all recursive calls by a unique one, namely, $\mathsf{change}_{\max_i(k_i)}^{o,o'}(X)$. In the end each variable is visited once, and the algorithm is indeed linear-time. □

Interestingly, a top-down algorithm is also possible: starting from the root nodes in the structure of $N$, inductively computes for each node $X$ the greatest value $k$ such that $N$ satisfies the formula $\mathsf{change}_k^{o,o'}(X)$. This algorithm allows us to derive the following result about *incomplete* deterministic CP-nets.

Say that a deterministic CP-net $N$ is *incomplete with a given structure* if it comes with a graph $G$ but for some variables $X$ and assignments $u$ to their parents, $N$ contains neither the rule $u{:}x{>}\bar{x}$ nor the opposite rule $u{:}\bar{x}{>}x$. Incomplete CP-nets arise naturally in the process of elicitation [Koriche and Zanuttini, 2009], and more generally when a user is indifferent to some objects (for instance: "I have no preferred colour for motorbikes, since I don't like motorbikes at all"). Then a *completion* of $N$ is a (complete, deterministic) CP-net with structure $G$ and containing the rules of $N$.

**Theorem 3.** *The problem of deciding whether there is at least one completion of a given, incomplete CP-net $N$ with a given tree structure, which entails $o{>}o'$ for given $o,o'$, can be solved in linear time $O(n)$.*

*Proof.* As evoked above, proceed top-down in the tree, by computing for each node $X$ the greatest $k$ for which there is a completion of $N$ satisfying $\mathsf{change}_k^{o,o'}(X)$. To do so, complete all missing rules in a greedy manner. For instance, if the current node $Y$ and its child $X$ are in the setting of Inductive Step 2 of Table 1, and $N$ contains no rule over $X$, choose the rules in the first disjunct to add in the completion of $N$. In this manner, from the value $k$ for $Y$ we get $k+1$ for $X$.

Obviously (because $\mathsf{change}_k^{o,o'}(X)$ reads "at least $k$ times"), the greater the value $k$ at each node, the more chances there are that the current completion indeed entails $o \succ o'$, hence the algorithm is correct. □

### 4.4 Hardness Result

We conclude this section by giving a hardness result, which sheds light on the difficulty of testing dominance in PCP-nets with a more general structure than a tree.

**Theorem 4.** *The problem of computing $p_\mathcal{N}(o \succ o')$, given a PCP-net $\mathcal{N}$ and two outcomes $o$ and $o'$, is #P-hard. This holds even if the structure is acyclic, the longest path has length 3, each node has at most one outgoing edge and at most 4 parents.*

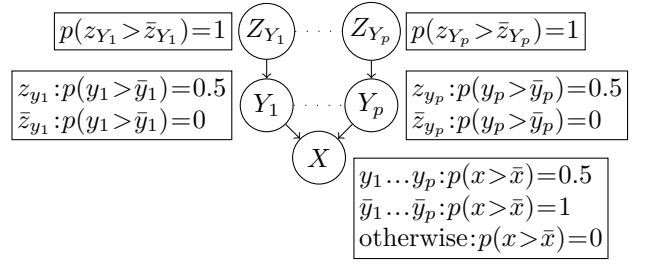

Figure 3: Reduction scheme

*Proof.* We give a reduction from #Monotone (2-4$\mu$) Bipartite CNF, which is #P-complete [Vadhan, 2002].

Let $\mathcal{X}$ and $\mathcal{Y}$ be two disjoint sets of variables. A monotone (2-4$\mu$)-bipartite CNF is a conjunction of clauses of the form $X \vee Y$, with $X \in \mathcal{X}$ and $Y \in \mathcal{Y}$, such that no variable appears more than 4 times in the formula. Given such a formula $\phi$, we build a PCP-net $\mathcal{N}$ over $\mathcal{V} = \mathcal{X} \cup \mathcal{Y} \cup \mathcal{Z}$, where $\mathcal{Z}$ contains one fresh variable, written $Z_Y$, for each $Y \in \mathcal{Y}$. The variables of $\mathcal{Z}$ have no parent, each $Y \in \mathcal{Y}$ has a single parent $Z_Y$, and each $X \in \mathcal{X}$ has for parents the $Y$'s such that the clause $X \vee Y$ appears in $\phi$ (there are at most 4 of them). This structure and the probability of each rule are given in Figure 3, where we show the portion of the PCP-net that corresponds to clauses $X \vee Y_1, \ldots, X \vee Y_p$.

Now consider the two outcomes $o, o'$ defined by $o[X] = x, o'[X] = \bar{x}$ for every $X \in \mathcal{X}$, $o[Y] = o'[Y] = y$ for every $Y \in \mathcal{Y}$, and $o[Z] = z, o'[Z] = \bar{z}$ for every $Z \in \mathcal{Z}$. We show that $p_\mathcal{N}(o \succ o')$ is exactly the proportion of interpretations of $\mathcal{V}$ in which $\phi$ is true.

Let $I$ be an interpretation of $\mathcal{X} \cup \mathcal{Y}$, and define the deterministic CP-net $N_I \propto \mathcal{N}$ as follows:

**(1) for every $\mathbf{Z} \in \mathcal{Z}$**, $N_I$ contains $z{>}\bar{z}$; and

**(2) for every $\mathbf{Y} \in \mathcal{Y}$:** (a) $N_I$ contains $\bar{z}_Y{:}\bar{y}{>}y$, and (b) if $I(Y) = \top$ then $N_I$ contains $z_Y{:}y{>}\bar{y}$, otherwise it contains the opposite rule

**(3) for every $\mathbf{X} \in \mathcal{X}$:** (a) $N_I$ contains $\bar{y}_1 \ldots \bar{y}_p{:}x{>}\bar{x}$, (b) if $I(X) = \top$ then $N_I$ contains $y_1 \ldots y_p{:}x{>}\bar{x}$, otherwise it contains the opposite rule, and (c) for all other assignments $u$ to $\mathsf{pa}(X)$, $N_I$ contains $u{:}\bar{x}{>}x$.

We show that $N_I$ entails $o{>}o'$ if and only if $I$ satisfies $\phi$. Clearly, we can reason on sets $\{X, Y_1, \ldots, Y_p, Z_{Y_1}, \ldots, Z_{Y_p}\}$ independently. So assume first that $I$ satisfies $(X \vee Y_1) \wedge \cdots \wedge (X \vee Y_p)$.

If $I$ satisfies $X$, then $I$ entails $o \succ o'$ using the worsening flips $z_{Y_1} > \bar{z}_{Y_1}, \ldots, z_{Y_p} > \bar{z}_{Y_p}$ and $y_1 \ldots y_p{:}x{>}\bar{x}$ (which can be performed in any order). Otherwise, $I$ must satisfy $Y_1 \wedge \cdots \wedge Y_p$, hence $I$ entails $o \succ o'$ using the flips $z_{Y_1}{:}y_1{>}\bar{y}_1, \ldots, z_{Y_p}{:}y_p{>}\bar{y}_p$, then the flip $\bar{y}_1 \ldots \bar{y}_p{:}x{>}\bar{x}$, then the flips $z_{Y_1} > \bar{z}_{Y_1}, \ldots, z_{Y_p} > \bar{z}_{Y_p}$, and finally the flips $\bar{z}_{Y_1}:\bar{y}_1 > y_1, \ldots, \bar{z}_{Y_p}:\bar{y}_p > y_p$.

The converse is shown similarly, and finally we have that $N_I$ entails $o > o'$ if and only if $I$ satisfies $\phi$. Now by construction, each $N_I$ built in this manner has a probability $1/2^n$ according to $\mathcal{N}$. Hence the probability with which $\mathcal{N}$ entails $o \succ o'$ is $m/2^n$ if and only if $\phi$ has $m$ models, which completes the reduction. □

## 5 Complexity of Optimisation

We now show that optimisation with tree-structured PCP-nets is both computationally easy and simple. The first result even holds for the much more general class of acyclic PCP-nets.

**Proposition 4.** *The probability for a given outcome $o$ to be optimal for a given acyclic PCP-net $\mathcal{N}$ can be computed in linear time $O(n)$.*

*Proof.* In the spirit of the "forward sweeping" procedure of Boutilier *et al.* [2004], it can be easily shown that $o$ is optimal for a deterministic CP-net $N \propto \mathcal{N}$ if and only if $N$ contains (1) the rule $o[X] > \bar{o}[X]$ for all root nodes $X$, and (2) the rule $o[\mathsf{pa}(X)]:o[X] > \bar{o}[X]$ for all other nodes $X$. It follows that the probability sought for is the product of the probabilities of all these rules, which can clearly be computed in time $O(n)$. □

**Proposition 5.** *The outcome with the maximal probability of being optimal for a given, tree-structured PCP-net $\mathcal{N}$ can be computed in linear time $O(n)$.*

*Proof.* The algorithm is a simple dynamic programming algorithm, operating bottom-up in the tree. First, given a leaf node $X$ with parent $Y$, the algorithm determines the optimal assignment to $X$ given $Y = y$, by taking the highest probability between rules $y:x > \bar{x}$ and $y:\bar{x} > x$, and similarly for $Y = \bar{y}$.

Now in the general case, given a variable $Y$ with parent $Z$ and children $X_1, \ldots, X_k$, the algorithm first considers the value $z$ for $Z$, and given this value searches for the most probable assignment to $Y, X_1, \ldots, X_k$ and their descendants. This can be done efficiently by comparing (1) $p_y \times p_{y1} \times \cdots \times p_{yk}$, where $p_y$ is the probability of the rule $z:y > \bar{y}$ and $p_{yi}$ ($i = 1, \ldots, k$) is the previously computed probability of the best assignment to $X_i$ and its descendants given $Y = y$, and (2) $p_{\bar{y}} \times p_{\bar{y}1} \times \cdots \times p_{\bar{y}k}$. Then the algorithm computes in a similar manner the probability of the most probable assignment given $Z = \bar{z}$, and based on this decides on the value $y$ or $\bar{y}$ for each of $z, \bar{z}$. Clearly, when all variables have been examined, the algorithm has computed the desired outcome. □

## 6 Conclusion

We proposed a "probabilistic" extension of conditional preference networks (CP-nets) for representing the preferences of a group of individuals over a set of combinatorial objects, or for representing ill-known preferences. We studied the probabilistic counterparts of the main reasoning tasks for CP-nets, namely dominance testing and optimisation, from the algorithmical and complexity viewpoints. We gave efficient algorithms for tree-structured probabilistic CP-nets, and as a by-product we obtained a linear-time algorithm for dominance testing in standard, tree-structured CP-nets.

As studied here, the expressiveness of our formalism is limited in two aspects. First, assuming a common, tree-like structure is unrealistic in some applicative settings. As future work, we plan to extend our results, in particular using a notion inspired from treewidth. The second limitation is due to the fact that the probability distribution on deterministic CP-nets which is represented by a probabilistic CP-net, is by definition an independent one (with rules as random variables). So as to allow PCP-net to model more realistic distributions, we plan to extend the representation by separating the probability distribution from the structure. An obvious choice is to use a Bayesian networks over the rules induced by the structure as random variables. Even with simple networks, this would allow, for instance, to represent fact such as: 3/4 of those individuals who prefer $x$ to $\bar{x}$ given $y$ also prefer $z$ to $\bar{z}$ given $t, u$. While one could fear a jump in complexity, it is worth noticing that our main result for tree-structured CP-nets goes through, in the sense that with such representation, computing the probability of $o \succ o'$ would amount to estimate the probability of $2^{2k^2}$ deterministic CP-nets, that is, to call an oracle for inference only a small number of times. This leaves hope that the framework can be extended to richer representations while preserving the low complexity of certain tasks.

## Acknowledgements

We thank Jérôme Lang and anonymous referees for their valuable comments on previous versions of this paper. The work presented here benefited of the support of the Agence Nationale de la Recherche, under grants ANR-11-BS02-008 and ANR-10-BLAN-0215.